\documentclass[10pt,twocolumn,letterpaper]{article}

\usepackage{graphicx}
\usepackage{booktabs}                   % Publication quality tables
\usepackage{multirow}
\usepackage{makecell}
\usepackage{paralist}
\usepackage[hyphens]{url}
\usepackage[pagebackref=true,breaklinks=true,colorlinks,bookmarks=false,linkcolor=blue,citecolor=blue]{hyperref}

\usepackage{amsmath}
\usepackage{amssymb}
\usepackage[accsupp]{axessibility}

\def\etal{et~al.}			  % and others, and co-workers
\newlength\paramargin
\newlength\figmargin
\newlength\secmargin
\newlength\figcapmargin
\newlength\tabcapmargin
\newlength\tabbotmargin

\newcommand{\Paragraph}[1]
{\vspace{1mm} \noindent\textbf{#1}}

\newcommand{\secref}[1]{Section~\ref{sec:#1}}
\newcommand{\figref}[1]{Figure~\ref{fig:#1}} 
\newcommand{\tabref}[1]{Table~\ref{tab:#1}}
\newcommand{\eqnref}[1]{\eqref{eq:#1}}

\long\def\ignorethis#1{}

\usepackage[pagenumbers]{cvpr} % To force page numbers, e.g. for an arXiv version

\def\cvprPaperID{1551} % *** Enter the CVPR Paper ID here
\def\confName{CVPR}
\def\confYear{2023}

\begin{document}

\title{
Consistent View Synthesis with Pose-Guided Diffusion Models
}

\author{
Hung-Yu Tseng$^1$\hspace{5mm}Qinbo Li$^1$\hspace{5mm}Changil Kim$^1$\hspace{5mm}Suhib Alsisan$^1$\hspace{5mm}Jia-Bin Huang$^{1,2}$\hspace{5mm}Johannes Kopf$^1$\vspace{0.5mm}\\
$^1$Meta\hspace{15mm}$^2$University of Maryland, College Park\vspace{0.5mm}\\
{\small\url{https://poseguided-diffusion.github.io/}}
}

\twocolumn[{
\renewcommand\twocolumn[1][]{#1}
\maketitle
\begin{center}
    \centering
    \vspace{-4mm}
    \includegraphics[width=0.95\linewidth]{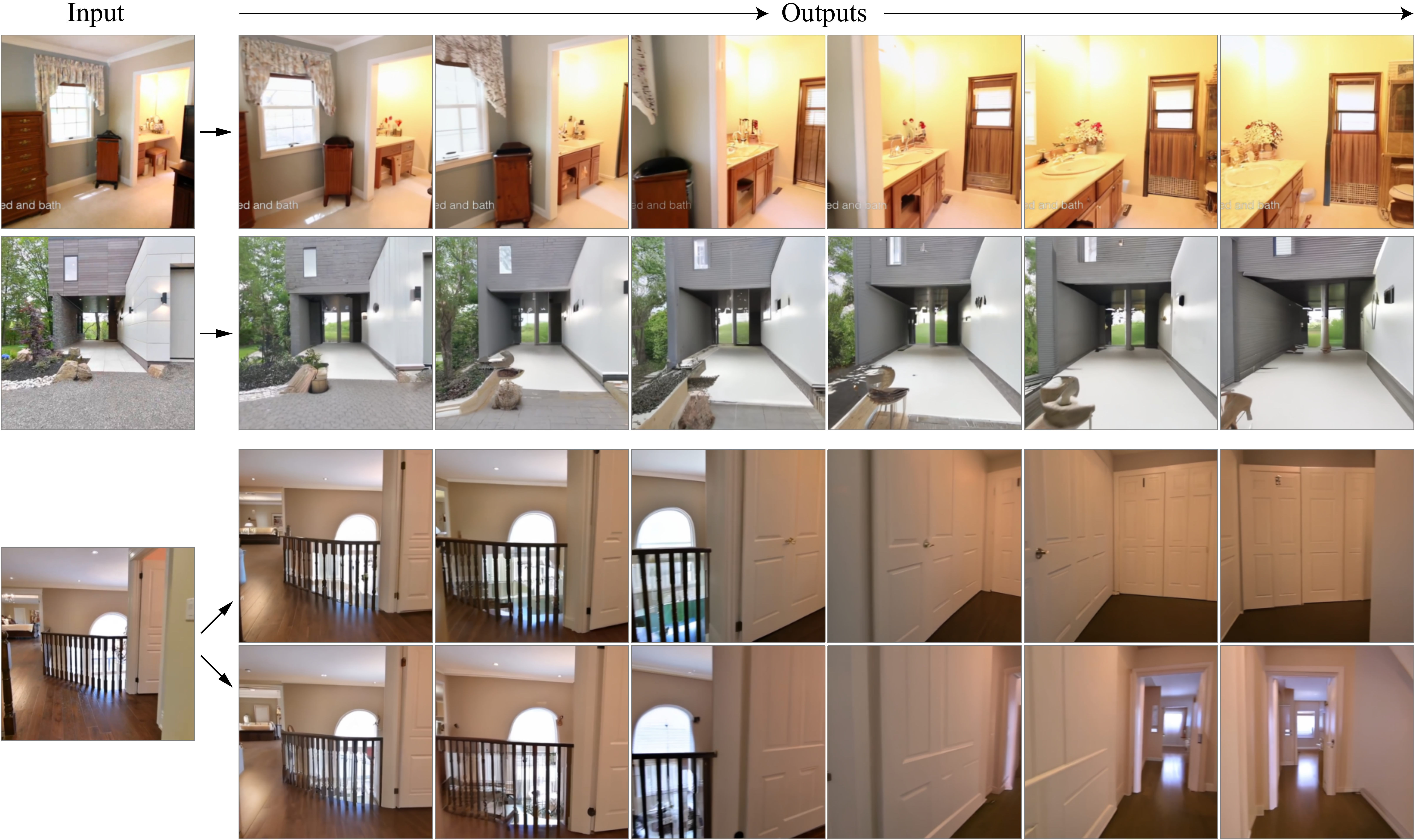}\captionsetup{type=figure}
    \vspace{-1.5mm}
    \captionof{figure}{
    \textbf{Consistent view synthesis via pose-guided diffusion model.} 
    (\textit{top}) Given an input image and a sequence of camera poses, we present a pose-guided diffusion model to generate a sequence of frames that are photorealistic and 3D consistent.
    (\textit{bottom}) Our proposed method can synthesize diverse sequences from the same set of inputs.}
    \label{fig:teaser}
\end{center}
}]

\begin{abstract}
Novel view synthesis from a single image has been a cornerstone problem for many Virtual Reality applications that provide immersive experiences.
However, most existing techniques can only synthesize novel views within a limited range of camera motion or fail to generate consistent and high-quality novel views under significant camera movement.
In this work, we propose a pose-guided diffusion model to generate a consistent long-term video of novel views from a single image.
We design an attention layer that uses epipolar lines as constraints to facilitate the association between different viewpoints.
Experimental results on synthetic and real-world datasets demonstrate the effectiveness of the proposed diffusion model against state-of-the-art transformer-based and GAN-based approaches.
\end{abstract}

\vspace{-4mm}
\section{Introduction}
\label{sec:intro}

Offering immersive 3D experiences from daily photos has attracted considerable attention.
It is a cornerstone technique for a wide range of applications such as 3D photo~\cite{kopf2020one,Shih3DP20}, 3D asset generation~\cite{dreamfusion}, and 3D scene navigation~\cite{bautista2022gaudi}.
Notably, rapid progress has been made in addressing the \textit{single-image view synthesis}~\cite{3dphoto,singleimagempi,synsin,pixelsynth} issue.
Given an arbitrarily narrow field-of-view image, these frameworks can produce high-quality images from novel viewpoints.
However, these methods are limited to viewpoints that are within a small range of the camera motion.

The \textit{long-term single-image view synthesis} task is recently proposed to address the limitation of small camera motion range.
As demonstrated in \figref{teaser}, the task attempts to generate a \emph{video} from a single image and a sequence of camera poses.
Note that different from the single-image view synthesis problem, the viewpoints of the last few video frames produced under this setting may be far away from the original viewpoint.
Take the results shown in \figref{teaser}, for instance, the cameras are moving into different rooms that were not observed in the input images.

Generating long-term view synthesis results from a single image is challenging for two main reasons.
First, due to the large range of the camera motion, e.g., moving into a new room, a massive amount of new content needs to be hallucinated for the regions that are not observed in the input image.
Second, the view synthesis results should be \emph{consistent} across viewpoints, particularly in the regions observed in the input viewpoint or previously hallucinated in the other views.

Both explicit- and implicit-based solutions are proposed to handle these issues.
Explicit-based approaches~\cite{pixelsynth,infinitenature,infinitenaturezero,se3ds} use a ``warp and refine" strategy.
Specifically, the image is first warped from the input to novel viewpoints according to some 3D priors, i.e., monocular depth estimation~\cite{midas,dpt}.
Then a transformer or GAN-based generative model is designed to refine the warped image.
However, the success of the explicit-based schemes hinges on the accuracy of the monocular depth estimation.
To address this limitation, Rombach~\etal~\cite{geogpt} designed a geometry-free transformer to implicitly learn the 3D correspondences between the input and output viewpoints.
Although reasonable new content is generated, the method fails to produce coherent results across viewpoints.
The LoR~\cite{lor} framework leverages the auto-regressive transformer to further improve the consistency.
Nevertheless, generating consistent, high-quality long-term view synthesis results remains challenging.

In this paper, we propose a framework based on diffusion models for consistent and realistic long-term novel view synthesis.
Diffusion models~\cite{sohl2015deep,song2019generative,ddpm} have achieved impressive performance on many content creation applications, such as image-to-image translation~\cite{saharia2022palette} and text-to-image generation~\cite{dalle2,imagen,balaji2022ediffi}.
However, these methods only work on 2D images and lack 3D controllability.
To this end, we develop a \emph{pose-guided} diffusion model with the epipolar attention layers. 
Specifically, in the UNet~\cite{ronneberger2015u} network of the proposed diffusion model, we design the epipolar attention layer to associate the input view and output view features.
According to the camera pose information, we estimate the epipolar line on the input view feature map for each pixel on the output view feature map.
Since these epipolar lines indicate the candidate correspondences, we use the lines as the constraint to compute the attention weight between the input and output views.

We conduct extensive quantitative and qualitative studies on real-world Realestate10K~\cite{re10k} and synthetic Matterport3D~\cite{mp3d} datasets to evaluate the proposed approach.
With the epipolar attention layer, our pose-guided diffusion model is capable of synthesizing long-term novel views that 1) have realistic new content in unseen regions and 2) are consistent with the other viewpoints.
We summarize the contributions as follows:
\begin{compactitem}
\item We propose a pose-guided diffusion model for the long-term single-image view synthesis task.
\item We consider the epipolar line as the constraint and design an epipolar attention to associate pixels in the images at input and output views for the UNet network in the diffusion model.
\item We validate that the proposed method synthesizes realistic and consistent long-term view synthesis results on the Realestate10K and Matterport3D datasets.
\end{compactitem}
\section{Related Work}
\label{sec:related}
\begin{figure*}[t]
\centering
\includegraphics[width=\linewidth]{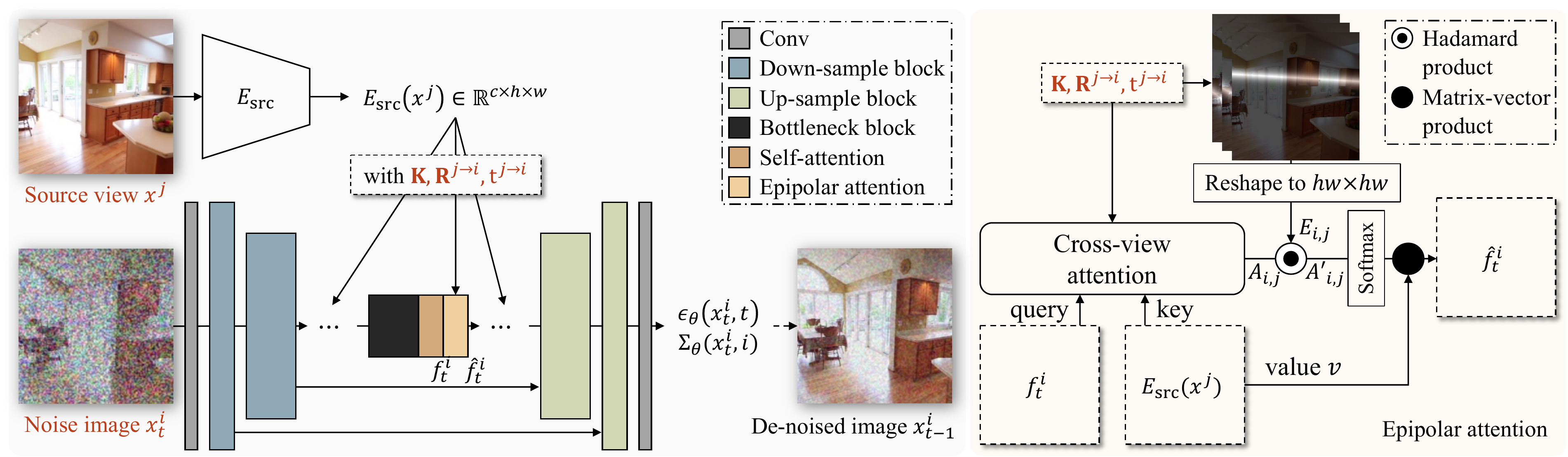}
\vspace{-5mm}
\caption{\textbf{Method overview.} 
(\textit{left}) The core component of our pose-guided diffusion model is the UNet that takes the source view image and camera poses as the input (red font), and de-noises the image at the target viewpoint.
We use an encoder to extract features from the source view features.
We design an epipolar attention to associate the target view with the source view features, and add the epipolar attention layer after \emph{each} self-attention layer in the UNet network.
The UNet model takes as input the source view features as well as the camera parameters via the epipolar attention layers, and predicts the de-noised target view image.
(\textit{right}) 
According to the input camera parameters, we compute the epipolar line as the constraint to estimate the attention between the source view and target view features.
}
\vspace{-2mm}
\label{fig:overview}
\end{figure*}

\Paragraph{Novel view synthesis.}
Novel view synthesis aims to generate high-quality images at arbitrary viewpoints given a set of posed images of a particular scene.
With the emergence of deep learning, early approaches~\cite{flynn2016deepstereo,tatarchenko2016multi,dosovitskiy2016learning} use Convolutional Neural Networks to synthesize novel views.
Instead of generating novel views directly, several methods~\cite{zhou2016view,srinivasan2017learning,park2017transformation,sun2018multi} predict the appearance flow for producing images of new viewpoints.
Recently, various 3D representations are leveraged for this task, including 3D point clouds~\cite{meshry2019neural,niklaus20193d,aliev2020neural,synsin,se3ds}, and layered representations such as layered depth images~\cite{kopf2020one,Shih3DP20} as well as multiplane images~\cite{zhou2018stereo,huang2020svs}.
These representations are used in 3D photo~\cite{kopf2020one}, light fields~\cite{mildenhall2019local,li2020synthesizing,attal2022learning} and many other novel view synthesis applications~\cite{srinivasan2019pushing,flynn2019deepview,singleimagempi}.
Very recently, neural radiance field (NeRF)~\cite{mildenhall2021nerf} methods reconstruct the target scene implicitly with multi-layer perceptrons and demonstrate impressive novel view synthesis results in various scenarios, including $360$-degree~\cite{zhang2020nerf++,barron2021mip} or large-scale~\cite{tancik2022block,meuleman2023localrf} 3D scenes.

Nevertheless, the approaches mentioned above can only 
1) \emph{interpolate} between multiple input views or 
2) \emph{extrapolate} from single/multiple views within a limited range of camera movement.
To synthesize a realistic novel view along camera trajectories that are far away from the input viewpoint, the PixelSynth~\cite{pixelsynth} and SE3DS~\cite{se3ds} schemes progressively construct a 3D point cloud from the input viewpoint according to the estimated depth, then repeatedly apply the ``warp and refine" strategy to produce novel views.
On the other hand, the GeoGPT~\cite{geogpt} framework uses a geometry-free transformer that does not rely on monocular depth estimation.
The LoR~\cite{lor} approach improves the transformer model to reduce the temporal flickering generated along a camera trajectory.
Nonetheless, it remains challenging to produce high-quality novel views.
In this work, we propose a pose-guided diffusion model that synthesizes a consistent and realistic sequence of novel views.

\Paragraph{Diffusion models.}
De-noising diffusion models~\cite{sohl2015deep,ddpm,song2019generative} are generative models that learn to generate data samples from Gaussian noise through a series of de-noising processes.
Recently, diffusion models have demonstrated remarkable performance on a variety of 2D content creation tasks, including image super-resolution~\cite{rombach2022high,wang2021s3rp,li2022srdiff,saharia2022image}, image in-painting~\cite{lugmayr2022repaint,saharia2022palette}, image de-blurring~\cite{lee2022progressive,whang2022deblurring}, and text-to-image~\cite{dalle2,imagen,balaji2022ediffi}.
In addition to working on 2D images, diffusion models are also emerging in the video generation~\cite{ho2022video,ho2022imagenvideo,voleti2022masked,singer2022make} or 3D shape generation~\cite{zhou20213d,luo2021diffusion,zeng2022lion} tasks.
As these methods lack 3D camera pose controllability, they cannot be directly applied to the view synthesis problem.
We also build upon diffusion models for our task. 
In contrast to existing diffusion models for image synthesis, our approach offers full controllability of the viewpoints.

Concurrent with our work, 3DiM~\cite{watson2022novel} also leverages diffusion models for view synthesis tasks.
Our work differs in two aspects.
First, 3DiM focuses on \emph{object-centric} synthetic scenes (e.g.,  ShapeNet dataset). 
In contrast, we focus on long-term view generation of \emph{scene-centric} realistic scenes with complex appearances.
Second, we exploit the epipolar constraints across views explicitly with the proposed epipolar cross-view attention layer. 
We demonstrate that integrating these geometric constraints leads to substantial quality improvements.

\Paragraph{Attention.}
Attention aims to capture the long-range dependencies, e.g., the relationship between two distant image pixels.
Attention mechanisms are widely used in deep learning tasks such as image recognition~\cite{liu2021swin} and image generation~\cite{zhang2019self}.
In particular, self-attention layers~\cite{vaswani2017attention,wang2018non} capture the dependencies within the \emph{same} data.
On the other hand, cross-attention~\cite{zhou2022cross} models the relationships between instances of \emph{different} data, e.g., two images, or an image vs. a text sequence.
The proposed epipolar attention can be considered as a type of cross-attention, where the epipolar lines are introduced as geometric constraints to compute the dependencies between the source view and target view image pixels.
\section{Methodology}
\label{sec:method}

Our goal is to synthesize a sequence of images $\{x^i\}^n_{i=2}$ given an input image $x^1$, and a sequence of camera poses $\{\mathbf{K}^i,\mathbf{R}^i, \mathbf{t}^i\}^n_{i=2}$, i.e., intrinsics, rotation, and translation, respectively.
We design a pose-guided diffusion model to auto-regressively generate the image at each viewpoint $i$ to produce the final sequence.
In this section, we first introduce diffusion models in \secref{3_1}.
We then illustrate the proposed pose-guided diffusion model in \secref{3_2}.
Finally, we describe how we produce the consistent novel view video in \secref{3_3}.

\subsection{Diffusion Model}
\label{sec:3_1}
Diffusion models~\cite{sohl2015deep,song2019generative,ddpm} learn to convert an empirical (i.e., isotropic Gaussian) distribution into the target data (i.e., real image) distribution through a series of de-noising operations.
A forward process is derived to gradually add noise to the real image so the image becomes indistinguishable from the Gaussian noise.
On the other hand, a backward process is learned to reverse the forward process, i.e., map from noises to real images.

\Paragraph{Forward and backward process.} Given an image $x_0$ sampled from the real image distribution $\mathcal{P}(x)$, the forward process converts the image to noise by a $T$-steps process that gradually adds Gaussian noise to $x_0$, namely
\begin{equation}
x_t=\sqrt{\alpha_t}x_{t-1} + (1-\alpha_t)\epsilon_t\hspace{5mm}\epsilon_t\sim\mathcal{N}(\mathbf{0},\mathbf{I}),
\end{equation}
where $t=[1,\cdots,T]$.
The notation $\alpha_t$ is computed from the noise schedule, which is pre-determined such that $x_T\approx\mathcal{N}(\mathbf{0}, \mathbf{I})$. 
We can further marginalize the forward process to
\begin{equation}
x_t=\sqrt{\bar{\alpha_t}}x_0 + (1-\bar{\alpha_t})\epsilon_t\hspace{5mm}\epsilon_t\sim\mathcal{N}(\mathbf{0},\mathbf{I}),
\end{equation}
where $\bar{\alpha_t}=\prod^t_{i=1}\alpha_i$.
The backward process can then be formulated as
\begin{equation}
\label{eq:backward}
x_{t-1}=\mu_\theta(x_t,t)+\Sigma_\theta(x_t,t)\epsilon_t\hspace{5mm}\epsilon_t\sim\mathcal{N}(\mathbf{0},\mathbf{I}),
\end{equation}
where $t=[T,\cdots,1]$.
Typically, an UNet~\cite{ronneberger2015u,dhariwal2021diffusion} model parameterized by $\theta$ is used to learn the backward process.

\Paragraph{Training.} We use the DDPM~\cite{ddpm} strategy that trains the UNet model to predict $\epsilon_\theta(x_t, t)$ instead of $\mu_\theta(x_t)$ in \eqnref{backward}, such that
\begin{equation}
\mu_\theta(x_t,t)=\big(x_t - (\frac{1-\alpha_t}{\sqrt{1-\bar{\alpha_t}}})\epsilon_\theta(x_t, t)\big)/\sqrt{\alpha_t}.
\end{equation}
The UNet model is trained using the mean square loss:
\begin{equation}
L_\mathrm{diffusion}=\mathbb{E}_{x,t}\big[\|\epsilon_t - \epsilon_\theta(x_t, t)\|^2_2\big].
\end{equation}
As for the term $\Sigma_\theta(x_t, t)$, we follow the improved DDPM~\cite{improvedddmp} approach that uses an additional objective $L_\mathrm{vlb}$ for training the UNet model to make the prediction.

\subsection{Pose-Guided Diffusion Model}
\label{sec:3_2}

\begin{figure}[t]
\centering
\includegraphics[width=0.95\linewidth]{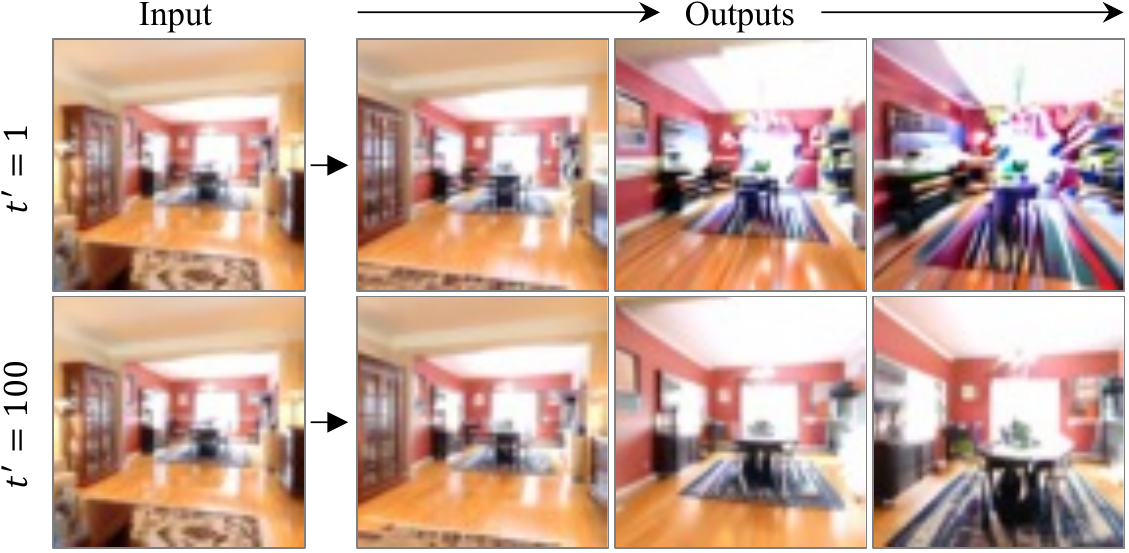}
\vspace{-2mm}
\caption{
\textbf{Artifacts from fixing noises in the backward process.} 
To improve the consistency across different views in the same sequence, we use the same set of initialization noise $x_T$ and diffusion noise $\{\epsilon_t\}^{t'}_{t=T}$ described in \eqnref{backward} to generate all views in the same video.
(\textit{top}) However, we find that fixing noises in all backward steps, i.e., $\{\epsilon_t\}^{1}_{t=T}$, creates obvious artifacts.
(\textit{bottom}) We address this by using the fixed noises in the early backward steps only, i.e., $\{\epsilon_t\}^{100}_{t=T}$, and re-sample the noises in the last few backward steps. This helps improve consistency while maintaining realism. 
}
\vspace{-2mm}
\label{fig:fixnoise}
\end{figure}
\begin{table*}[t]
    \caption{\textbf{Quantitative evaluation on short-term view synthesis.} We report the average PSNR ($\uparrow$), SSIM ($\uparrow$), and LPIPS ($\downarrow$) scores between the first five generated and ground-truth frames in the videos. The best performance is in \textbf{bold}.}
    \vspace{-2mm}
    \label{tab:short_term}
    \centering
    \footnotesize
    \begin{tabular}{l|ccc ccc} 
	    \toprule
		\multirow{2}{*}{Methods} &  \multicolumn{3}{c}{Re10K} & \multicolumn{3}{c}{MP3D} \\
		\cmidrule(lr){2-4} \cmidrule(lr){5-7} & PSNR ($\uparrow$) & SSIM ($\uparrow$) & LPIPS ($\downarrow$) & PSNR ($\uparrow$) & SSIM ($\uparrow$) & LPIPS ($\downarrow$) \\
		\midrule
		GeoGPT~\cite{geogpt} & $20.90$ & $0.61$ & $2.53$ & $16.87$ & $\mathbf{0.63}$ & $3.46$ \\
		LoR~\cite{lor} & $20.93$ & $0.61$ & $2.35$ & $19.64$ & $0.61$ & $3.30$\\
		SE3DS~\cite{se3ds} & $18.24$ & $0.59$ & $3.20$ & - & - & - \\
		Ours & $\mathbf{22.64}$ & $\mathbf{0.68}$ & $\mathbf{2.19}$ & $\mathbf{20.59}$ & $\mathbf{0.63}$ & $\mathbf{2.90}$ \\
		\bottomrule
    \end{tabular}
    \vspace{-2mm}
\end{table*}

% not fix noise
% 22.79, 0.69, 2.18
% fix noise
% 22.64, 0.68, 2.19
\begin{figure*}[t]
\centering
\includegraphics[width=\linewidth]{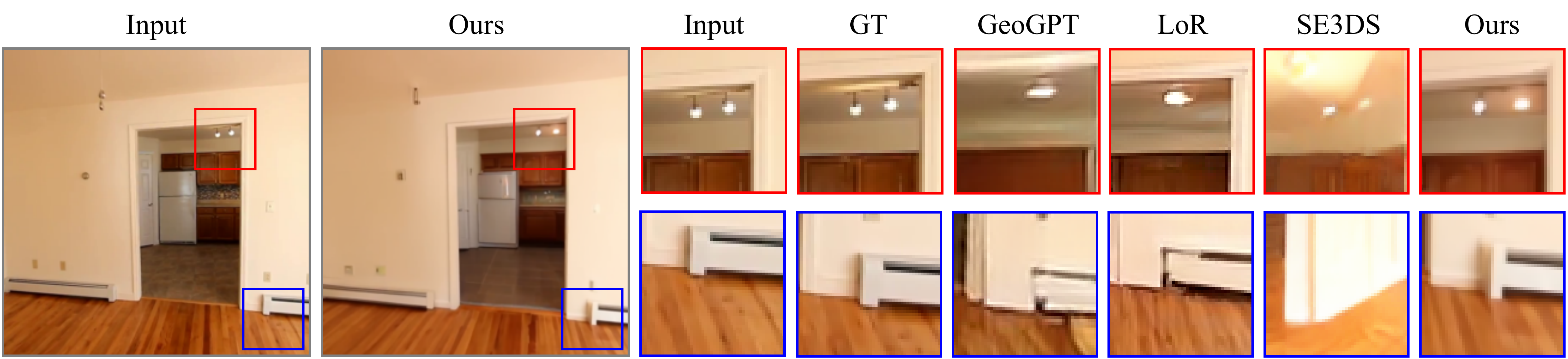}
\vspace{-5mm}
\caption{\textbf{Qualitative comparisons of short-term view synthesis.} 
We present the short-term single-image view synthesis results generated by different methods. The patches are all cropped from the same location of the patches in the second image (i.e., Ours).
}
\vspace{-2mm}
\label{fig:shortterm}
\end{figure*}

We present an overview of the proposed pose-guided diffusion model in \figref{overview}.
Given the source view image $x^j$ at the $j$-th viewpoint, where $j\in [1,\cdots,n]$, the goal is to de-noise the target view image $x^i_t$ at the diffusion time step $t$.
We first use the source view encoder to extract the feature maps from the source view image $x^j$.
Combining the feature maps using the proposed epipolar attention layer, the UNet model predicts the $\epsilon_\theta(x^i_t, t)$ and $\Sigma_\theta(x^i_t, t)$ terms to estimate the de-noised image $x^i_{t-1}$.
We obtain the final target view image $x^i_0$ by iterating through the backward process.

\Paragraph{Source view encoder.}
Given the source view image $x^j$, we use a deep convolutional neural network $E_\mathrm{src}$ to extract the feature map $E_\mathrm{src}(x^j)\in\mathbb{R}^{c\times h\times w}$, where $c\times w$ matches the resolution of the attention layer in the UNet network.
In practice, we use the pre-trained MiDaS~\cite{midas} model as the source view encoder.
Our early experiments show that such a strategy facilitates faster training of the pose-guided diffusion model.
Note that we extract multiple intermediate feature maps from the MiDas model according to the resolutions of the attention layers used in the UNet model.

\Paragraph{UNet network.}
We modify the commonly-used UNet architecture in diffusion models~\cite{dhariwal2021diffusion} as our UNet network.
As demonstrated in the left part of \figref{overview}, we add the proposed epipolar attention layer after \emph{each} of the self-attention layers in the UNet network.

\Paragraph{Epipolar attention.}
The proposed epipolar attention aims to \emph{associate} the target view with the source view.
The core idea is to leverage the epipolar line as the \emph{constraint} to reduce the number of candidate source view pixels corresponding to a particular target view pixel.
We present the epipolar attention in the right-hand side of \figref{overview}.
Given the query calculated from intermediate UNet feature $f^i_t$ and the key computed from the source view feature $E_\mathrm{src}(x^j)$, we first use the cross-view attention~\cite{zhou2022cross} to compute the affinity matrix $A_{i,j}\in\mathbb{R}^{hw\times hw}$.
The term $h\times w$ indicates the resolution of the epipolar attention layer.
Second, for each pixel position on the intermediate UNet feature map $f^i_t$, we compute the epipolar line on the source view feature map $E_\mathrm{src}$ according to the camera parameters $\mathbf{K}$, $\mathbf{R}^{j\rightarrow i}$, and $\mathbf{t}^{j\rightarrow i}$.
The line is then converted to a weight map of shape $h\times w$ where the values indicate the inverse distance to the epipolar line.
We estimate the weight maps for all positions in $f^i_t$, stack these maps, and reshape to get the epipolar weight matrix $E_{i,j}\in\mathbb{R}^{hw\times hw}$.
We re-weight the affinity matrix by $A'_{i,j}=A_{i,j}\odot E_{i,j}$, where $\odot$ denotes the Hadamard product.
Finally, the output of the epipolar attention layer $\hat{f}^i_t\in\mathbb{R}^{c\times h\times w}$ is computed as
\begin{equation}
\label{eq:epipolar}
\hat{f}^i_t=\mathrm{reshape}\big(\mathrm{softmax}(A'_{i,j})\cdot v\big),
\end{equation}
where $v$ is the value term calculated from the source view feature map $E_\mathrm{src}(x^j)$.
We detail the computation of the epipolar line in \secref{appendix_tech_details}.

\Paragraph{Super-resolution.}
We use the cascaded diffusion~\cite{ho2022cascaded,dalle2,imagen} strategy to obtain the final spatial resolution.
Specifically, we use a base pose-guided diffusion model to produce the sequence of resolution $64\times 64$.
Then another pose-guided super-resolution diffusion model, detailed in \secref{appendix_tech_details}, is used to generate the final $256\times 256$ video.

\subsection{Consistent Long-Term View Synthesis}
\label{sec:3_3}
Our goal is to synthesize \emph{a sequence of} novel views given the input image.
Although the proposed pose-guided diffusion model learns to generate a single novel view during the training time, we can use the auto-regressive inference to produce long-term view synthesis in the test time.
A simple way is to consider the target view $x^{i-1}$ generated at the previous step as the source view $x^j$ to generate the novel view in the current step, i.e., $j=i-1$.
Nevertheless, this approach produces temporal flickering in the final video due to the frame-by-frame processing strategy.
We use the following two solutions to address the issue.

\Paragraph{Stochastic conditioning.}
We find that using stochastic conditioning~\cite{watson2022novel} slightly improves the temporal flickering.
Specifically, at each step in the backward process described in \eqnref{backward}, instead of using the previous frame $x^{i-1}$, we randomly sample the source view image $x^j$ from the set of prior frames $x^j\sim\mathrm{Uniform}(\{x^k,\cdots,x^{i-1}\})$.
Such a strategy encourages the diffusion model to be guided by all the previous frames, thus improving the temporal consistency.

\Paragraph{Fixing noises in the backward process.}
The noises introduced during the backward process illustrated in \eqnref{backward} also contribute to the temporal inconsistency.
To reduce the variance of the backward process across different views, we use the same initialization noise $x_T$ and diffusion noises $\{\epsilon_t\}^1_{t=T}$ to generate all images in the same video.
Nevertheless, we observe noticeable artifacts if we fix all diffusion noises $\{\epsilon_t\}^1_{t=T}$ during the backward process, as demonstrated in \figref{fixnoise}.
In practice, fixing the diffusion noises $\{\epsilon_t\}^{t'}_{t=T}$ to a certain backward step $t'$ alleviates the issue and improves the temporal consistency.\footnote{We set $t'$ to be 100 in all experiments, indicating that we re-sample the noise $\epsilon$ in the last $100$ backward steps.}
\section{Experimental Results}
\label{sec:exp}

\begin{figure*}[t]
\centering
\includegraphics[width=0.9\linewidth]{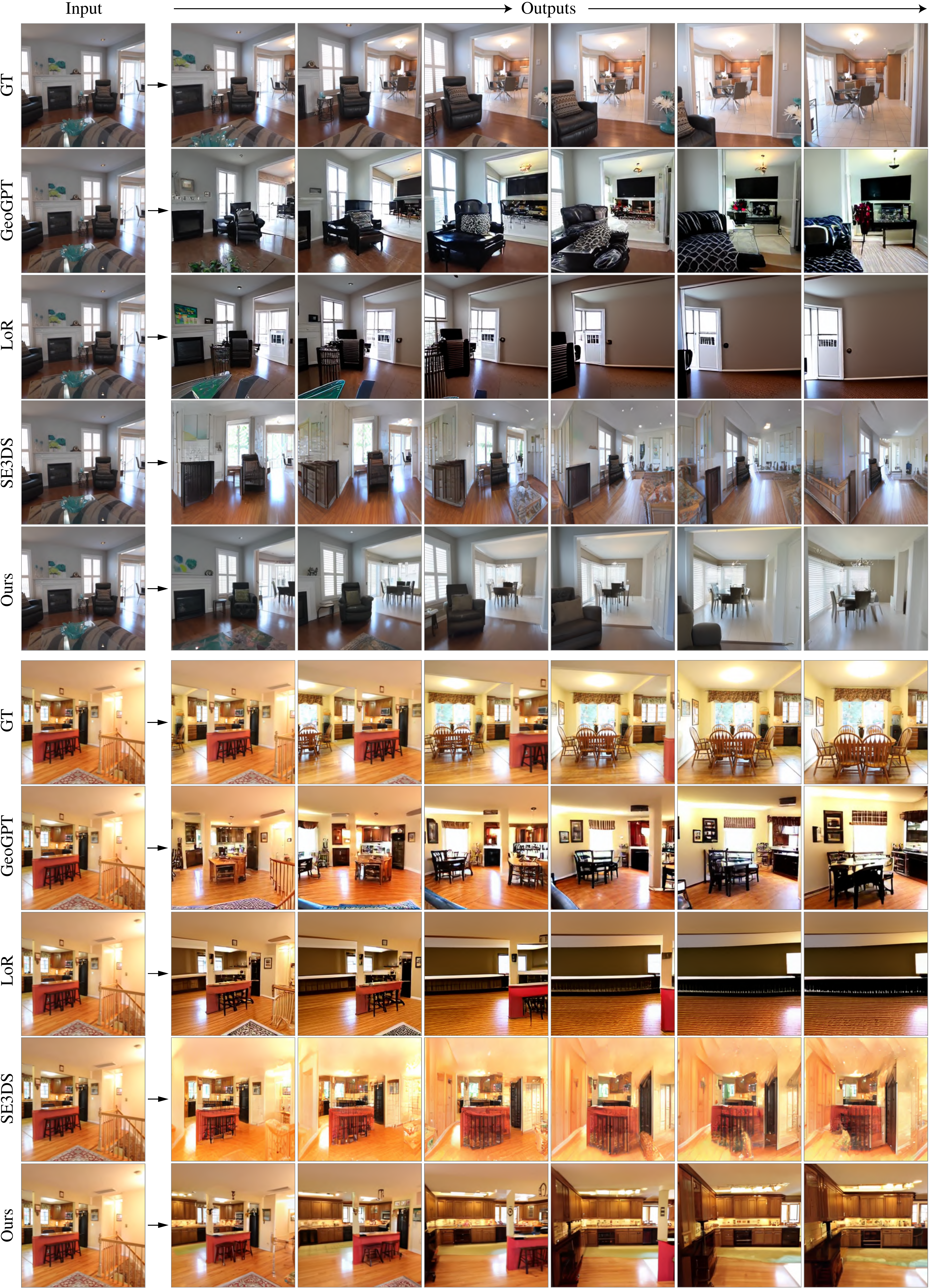}
\vspace{-3mm}
\caption{\textbf{Qualitative comparisons.} We present the long-term single-image view synthesis results generated by different methods. 
}
\label{fig:qualitative}
\end{figure*}

\subsection{Experimental Setup}

\Paragraph{Datasets.}
We focus on two multi-view datasets for all experiments: real-world RealEstate10K (Re10K)~\cite{re10k} and synthetic Matterport 3D (MP3D)~\cite{mp3d}.
We use $61,986$ video clips in the Re10K dataset for training and randomly sample $500$ sequences from the testing split for the evaluation.
As for the MP3D dataset, we follow the common protocol~\cite{synsin,pixelsynth,videoautoencoder,lor} to use the Habitat agent~\cite{habitat} to render $6,000$ training videos and $500$ testing videos.
For both datasets, we resize and center-crop the video to the spatial resolution of $256\times 256$.

\Paragraph{Compared methods.}
We compare our method with several state-of-the-art methods: two recent transformer-based approaches GeoGPT~\cite{geogpt} and LoR~\cite{lor}, as well as a very recent GAN-based scheme SE3DS~\cite{se3ds}.

\Paragraph{Evaluation setting.}
We evaluate the short-term and long-term view synthesis results. 
We generate a $20$-frames video for each testing image, and consider the first $5$ frames as the short-term views:
\begin{compactitem}
\item Short-term: We use pairwise metrics PSNR, SSIM, and LPIPS~\cite{lpips} to measure the difference between the generated and ground-truth images.
\item Long-term: We measure generated image quality and temporal consistency.
For image quality, we use the FID~\cite{fid} and KID~\cite{kid} scores to estimate the realism of the last (i.e., $20$-th) generated frame.
We use the flow warping error ($E_\mathrm{warp}$)~\cite{flowwarpingerror} to quantify the temporal consistency.
Specifically, we use the RAFT~\cite{teed2020raft} model to compute optical flow between two consecutive generated frames.
Then the error is computed as
\begin{equation}
E_\mathrm{warp} = \sum\limits_{i=2}^{20} M^{(i-1)\rightarrow i} \lVert x^i - \hat{x}^{i-1}\rVert_1,
\end{equation}
where $M$ is the visibility mask, and $\hat{x}^{i-1}$ is warped from the output frame $x^{i-1}$ using to the optical flow.
\end{compactitem}
More details are provided in \secref{appendix_exp_details}.

\subsection{Short-term View Synthesis}
We present the quantitative comparisons in \tabref{short_term} and qualitative results in \figref{shortterm}.
While the SE3DS method struggles to produce realistic results, the GeoGPT and LoR frameworks have similar performance on producing short-term novel views.
However, the details generated by these two transformer-based methods are slightly inconsistent with the input view.
In contrast, the proposed approach synthesizes 1) details that are consistent with the input view and 2) accurate parallax that corresponds to the camera motion.

\subsection{Long-Term View Synthesis}
\begin{table*}[t]
    \caption{\textbf{Quantitative evaluation on long-term view synthesis.} Given the $20$-frames videos, we report the average FID ($\downarrow$) and KID ($\downarrow$) scores of the last generated frames, and use all generated frames to compute the flow warping error $E_\mathrm{warp}$ ($\downarrow$). The best performance is in \textbf{bold}. We also report the score of real testing videos for reference.}
    \vspace{-2mm}
    \label{tab:long_term}
    \centering
    \footnotesize
    \begin{tabular}{l|ccc ccc} 
	    \toprule
		\multirow{2}{*}{Methods} &  \multicolumn{3}{c}{Re10K} & \multicolumn{3}{c}{MP3D} \\
		\cmidrule(lr){2-4} \cmidrule(lr){5-7} & FID ($\downarrow$) & KID ($\downarrow$) & $E_\mathrm{warp}$ ($\downarrow$) & FID ($\downarrow$) & KID ($\downarrow$) & $E_\mathrm{warp}$ ($\downarrow$) \\
		\midrule
		Real & $41.09$ & $0.011$ & $0.018$ & $58.83$ & $0.011$ & $0.019$ \\
		\midrule
		GeoGPT~\cite{geogpt} & $63.30$ & $\mathbf{0.016}$ & $0.073$ & $213.14$ & $0.046$ & $0.057$  \\
		LoR~\cite{lor} & $98.01$ & $0.034$ & $0.030$ & $113.50$ & $0.048$  & $0.036$  \\
		SE3DS~\cite{se3ds} & $235.8$ & $0.153$ & $0.060$ & - & - & - \\
		Ours & $\mathbf{56.33}$ & $\mathbf{0.016}$ & $\mathbf{0.023}$ & $\mathbf{72.48}$ & $\mathbf{0.019}$ & $\mathbf{0.035}$ \\
		\bottomrule
    \end{tabular}
    \vspace{-3mm}
\end{table*}

% not fix noise
% 56.57, 0.019, 0.026
% fix noise
% 56.33, 0.016, 0.023
We measure the last frame FID and KID scores to evaluate the per-frame quality, and calculate the flow warping error $E_\mathrm{warp}$ to access the temporal consistency of the generated $20$-frames videos.
We demonstrate the quantitative comparisons in \tabref{long_term}, and show example qualitative results in \figref{qualitative}.
Similar to the short-term view synthesis setting, the SE3DS scheme struggles to generate appealing results, especially under large camera motion, e.g., the bottom example in \figref{qualitative}.
On the other hand, the GeoGPT model synthesizes realistic novel views.
Nevertheless, the results are not consistent across different viewpoints, i.e., the scene changes drastically frame-by-frame.
In contrast to the GeoGPT approach, the novel views produced by the LoR method are more consistent.
Nonetheless, we observe a quality degradation in the last few generated frames.
Compared to these existing approaches, our model generates novel view sequences that 1) maintain the image quality over time and 2) contain less temporal flickering.

\begin{figure}[t]
\centering
\includegraphics[width=0.85\linewidth]{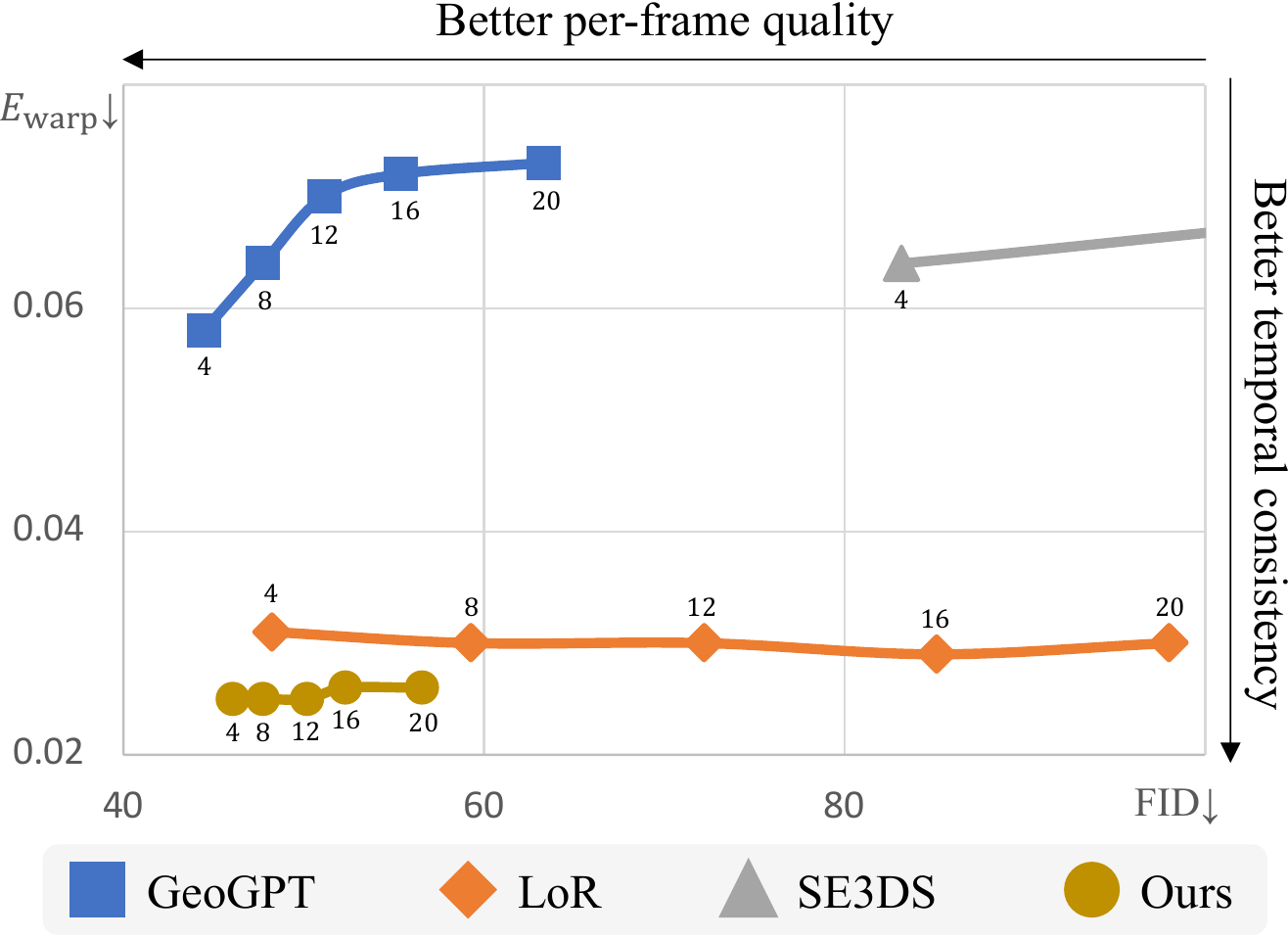}
\vspace{-2mm}
\caption{
\textbf{Last frame quality vs. temporal flickering.} 
We show the FID ($\downarrow$) of the last frames and flow-warping errors $E_\mathrm{warp}$ ($\downarrow$) given different generated video lengths $\{4,8,\cdots,20\}$.
Our method generates not only realistic but also consistent long-term single-image view synthesis results.
}
\vspace{-2mm}
\label{fig:fidwarp}
\end{figure}

% Last frame FID
%  4 | 8 | 12 | 16 | 20
% GeoGPT: 44.48 | 47.74 | 51.14 | 55.38 | 63.3
% LoR: 48.24 | 59.26 | 72.21 | 85.12 | 98.01
% SE3DS: 83.19 | 117.43 | 160.38 | 210.75 | 235.84
% Ours: 46.08 | 47.73 | 50.19 | 52.30 | 56.57

% Average Flow Warping Error
%  4 | 8 | 12 | 16 | 20
% GeoGPT: 0.058 | 0.064 | 0.070 | 0.072 | 0.073 
% LoR: 0.031 | 0.030 | 0.030 | 0.029 | 0.030
% SE3DS: 0.064 | 0.069 | 0.071 | 0.065 | 0.060
% Ours: 0.025 | 0.025 | 0.025 | 0.026 | 0.026

\Paragraph{Per-frame quality vs. temporal consistency.}
It is challenging to assess the overall long-term view synthesis performance since there are two perspectives: per-frame quality (FID, KID) and temporal consistency ($E_\mathrm{warp}$).
Therefore, we plot the FID vs. $E_\mathrm{warp}$ curves of videos with different lengths generated by various methods in \figref{fidwarp}.
Consistent with the observation we have from \tabref{long_term} and \figref{qualitative}, the GeoGPT model fails to generate consistent images, while the LoR approach struggles to maintain the generated image quality over time.
In contrast, the proposed pose-guided diffusion model synthesizes novel views that are consistent and remain realistic over time.

\subsection{Ablation Study}
We conduct ablation studies using the Re10K dataset to further analyze the proposed approach.

\Paragraph{Epipolar attention.}
In order to understand the effectiveness of the proposed epipolar attention, we make a comparison to two baselines: Concat and Cross-view attention.
In Concat baseline, we use the commonly-used UNet~\cite{dhariwal2021diffusion} structure with two modifications.
First, we concatenate the source view $x^j$ with the noise image $x^i_t$ at the target view as the input to the UNet model.
Second, we flatten the input camera pose parameters, compute the embedding vector and add the vector to the diffusion time-step embedding for the UNet network.\footnote{The strategy is similar to adding class conditioning embedding~\cite{dhariwal2021diffusion}, or adding text embedding~\cite{imagen} to the diffusion time-step embedding.}
As for the Cross-view attention baseline, we simply remove the epipolar constraint (i.e., use $A_{i,j}$ instead of $A'_{i,j}$ in \eqnref{epipolar}) in the proposed epipolar attention layer.
To ensure a fair comparison, we use identical hyper-parameters to train different models to generate $64\times 64$ sequences, then use a third-party video super-resolution~\cite{realbasicvsr} model to get the final results of resolution $256\times 256$.
Furthermore, all the compared methods use stochastic conditioning and noise-fixing.

We present the results in \tabref{epipolar}.
Compared to the Cross-view attention baseline, the Concat baseline fails to generate high-quality novel views in long-term, since it is challenging to learn the correspondence between source and target views via concatenated inputs. 
On the other hand, our approach synthesizes realistic novel views as the proposed attention leverages epipolar lines as the constraint to estimate the dependency between the source and target views.

\begin{table}[t]
    \caption{\textbf{Impact of epipolar attention.} We report the FID ($\downarrow$) and KID ($\downarrow$) scores of the last generated video frames. We use different diffusion models to generate the $64\times64$ sequences, then use the same video super-resolution~\cite{realbasicvsr} model to get the $256\times256$ videos for fair comparison. The best performance is in \textbf{bold}.}
    \vspace{-2mm}
    \label{tab:epipolar}
    \centering
    \footnotesize
    \begin{tabular}{l|cc} 
	    \toprule
		\multirow{2}{*}{Methods} &  \multicolumn{2}{c}{Re10K} \\
		\cmidrule(lr){2-3} & FID ($\downarrow$) & KID ($\downarrow$) \\
		\midrule
		Source/target views concatenation & $87.22$ & $0.034$\\
		Cross-view attention & $81.37$ & $0.033$ \\
		Epipolar attention (Ours) & $\mathbf{69.63}$ & $\mathbf{0.025}$\\
		\bottomrule
    \end{tabular}
    \vspace{-3mm}
\end{table}

% PSNR/SSIM/LPIPS/KID/FID/WarpError
% --- VSR --- 
% concat: 23.43, 0.71, 2.10, 87.22, 0.034, 0.014
% cv    : 23.05, 0.70, 2.22, 81.37, 0.033, 0.014
% ours  : 22.78, 0.69, 2.28, 69.63, 0.025, 0.014

% ours w/ fixing noise: 22.63, 0.69, 2.28, 73.95, 0.026

% ours 64: 25.83, 0.79, 1.09

\Paragraph{Super-resolution.}
In this study, we compare different super-resolution approaches: monocular image super-resolution (ESRGAN)~\cite{realesrgan}, video super-resolution (RealBasicVSR)~\cite{realbasicvsr}, and our pose-guided super-resolution diffusion model.
For a fair comparison, we use the same $64\times 64$ sequences generated by the low-resolution pose-guided diffusion model as the input.
The results are shown in \tabref{super_res}.
The videos super-resolved by the RealBasicVSR method contain less flickering compared to the other methods since they process the low-resolution sequences frame-by-frame.
On the other hand, the pose-guided diffusion model generates much more high-quality novel views.
Therefore, we use the pose-guided diffusion model to super-resolve the low-resolution novel view videos in all experiments.
Nevertheless, we argue that video super-resolution diffusion models may be critical to further reduce the temporal flickering while maintaining the visual quality.

\begin{table}[t]
    \caption{\textbf{Super-resolution models.} 
    We report the average LPIPS ($\downarrow$) scores for the short-term, FID ($\downarrow$) and $E_\mathrm{warp}$ scores for the long-term novel view synthesis results. We use the same $64\times64$ results and different super-resolution methods to get the $256\times256$ videos. The best performance is in \textbf{bold}.}
    \vspace{-2mm}
    \label{tab:super_res}
    \centering
    \footnotesize
    \begin{tabular}{l|ccc} 
	    \toprule
		\multirow{2}{*}{Methods} &  \multicolumn{3}{c}{Re10K} \\
		\cmidrule(lr){2-4} & LPIPS ($\downarrow$) & FID ($\downarrow$) & $E_\mathrm{warp}$ ($\downarrow$)\\
		\midrule
		Real-ESRGAN~\cite{realesrgan} & $2.32$ & $75.05$ & $0.021$ \\
            RealBasicVSR~\cite{realbasicvsr} & $2.28$ & $69.63$ & $\mathbf{0.014}$\\
		Pose-guided diffusion model & $\mathbf{2.19}$ & $\mathbf{56.33}$ & $0.023$ \\
		\bottomrule
    \end{tabular}
    \vspace{-1mm}
\end{table}

\vspace{-1mm}
\section{Limitations and Future Works}
\vspace{-1mm}
\begin{figure}[t]
\centering
\includegraphics[width=0.85\linewidth]{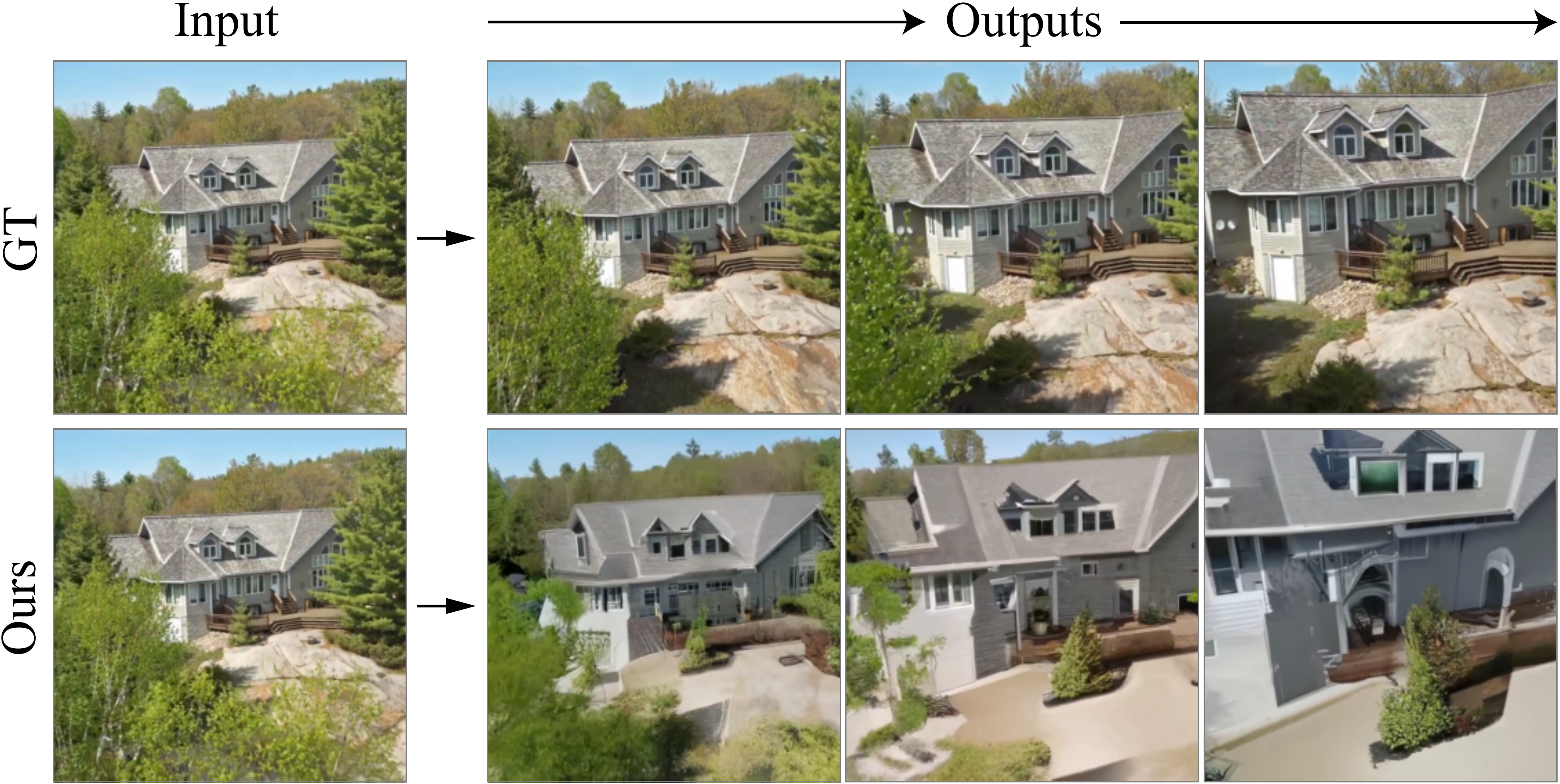}
\vspace{-2mm}
\caption{\textbf{Failure case.} 
Our proposed method fails to generate realistic novel views if the scale of the scene is significantly different from those in the training data.
} 
\vspace{-3.5mm}
\label{fig:failure}
\end{figure}
The proposed method has the following limitations.
First, our approach cannot handle the case where scene scales vary dramatically across different videos, e.g., landscape videos explored in \cite{infinitenature,infinitenaturezero}.
Take \figref{failure}, for instance, the scale of the scene is significantly larger than those in the Re10K training data.
We believe that handling such cases requires proper scale normalization or data augmentation. 
We leave this exploration to future work.
Second, the inference is time-consuming as it involves multiple steps (i.e., $250$ in practice) in the backward process to predict one single novel view.
As many recent efforts~\cite{ddim,salimans2021progressive} are made to accelerate the inference speed of the diffusion model, we plan to explore these solutions in the future.
\vspace{-1.5mm}
\section{Conclusions}
\vspace{-1mm}
In this work, we introduce a pose-guided diffusion model to synthesize a novel view video under massive camera motion from a single image.
The core of our diffusion model is the epipolar attention that estimates the dependencies between images of two camera viewpoints.
Qualitative and quantitative results show that the proposed pose-guided diffusion model generates novel views that are 1) realistic, even the viewpoints far away from the input view, and 2) consistent across various viewpoints.
%\pagebreak

%%%%%%%%% REFERENCES
{\small
\bibliographystyle{ieee_fullname}
\bibliography{egbib}
}

\clearpage
\onecolumn
\appendix
\section{Supplementary Materials}

% --- overview --- %
\subsection{Overview}

In this supplementary material, we first present qualitative results of the existing and our methods.
Second, we provide the technical details of the proposed approach to complement the paper.
Finally, we describe the details of the experiments, including the dataset pre-processing, evaluation metric computation, and reproducing existing approaches.

\subsection{Qualitative Results}
%We provide qualitative results, i.e., \emph{videos}, in the attached HTML webpage.
%
%Please use the file \href{./main.html}{main.html} to visualize the following results.
We provide more qualitative results, i.e., \emph{videos} on the project page.\footnote{\url{https://poseguided-diffusion.github.io}}

\Paragraph{Comparing with previous SOTA methods.} We show the single-image view synthesis results of difference scenes generated by the GeoGPT~\cite{geogpt}, LoR~\cite{lor}, SE3DS~\cite{se3ds} and the proposed pose-guided diffusion models.

\Paragraph{Generating diverse outputs}: We demonstrate that the proposed method is capable of producing multiple realistic videos from the same set of inputs.

\Paragraph{Ablation study}: We show the qualitative results generated by the Source/target views concatenation and Cross-view attention baselines described in Table 3 in the paper.
Specifically, we use different diffusion models to generate the $64\times 64$ videos.
We then use the \emph{same} super-resolution diffusion model introduced in \secref{3_2} in the paper to get the final $256\times 256$ results.
Consistent with the quantitative results shown in Table 3 in the paper, the per-frame quality of the videos generated by the baseline approaches degrades significantly in the final few frames.

\Paragraph{Long-range view interpolation}: In addition to single-image view synthesis, the proposed method can also \emph{interpolate between two far-away viewpoints}.
The key is to leverage the stochastic conditioning approach described in \secref{3_3} of the paper.
Specifically, to interpolate the view $i$ between two viewpoints $l$ and $k$, we randomly sample the source view image $x^j$ from the two viewpoints, i.e., $x^j\sim P(\{x^l, x^k\})$ during the backward process in the diffusion model. 
The sampling probabilities of each input view are determined inverse proportional to the distance to the output viewpoint.
In this experiment, we take the $1$st, $11$th, and $21$th frames from each testing video as the input.
We interpolate between these input views to obtain a $21$-frames video as the result.
As no existing approach tackles the long-range view interpolation problem, we compare our approach with an alternative large-motion frame interpolation approach FiLM~\cite{reda2022film}.

\Paragraph{Failuer cases.} We show several example failure cases.
First, the proposed method cannot candle the case where the scene scale is significantly different to those in the training data, e.g., natural scenes.
Second, our approach fails to produce high-quality results with rare camera pose sequences, e.g., going up/down stairs.

\begin{figure}[t]
\centering
\includegraphics[width=0.75\linewidth]{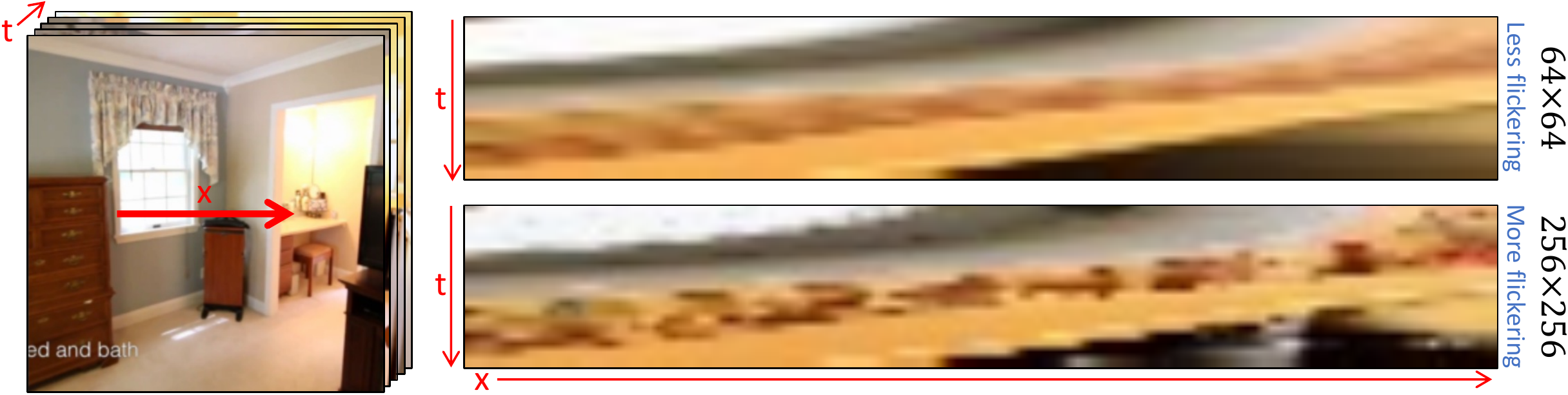}
\caption{\textbf{X-T slice visualization.}
We visualize the x-t slice of the low-resolution (\textit{top}) and high-resolution (\textit{bottom}) videos.
The flickering mainly comes from the super-resolution stage.
}
\label{fig:xtslice}
\end{figure}

\Paragraph{Flickering caused by super-resolution.} 
To understand the flickering produced by our algorithm, we present the x-t slice visualization in \figref{xtslice}.
The flickering mainly comes from the per-frame super-resolution stage.

\subsection{Technical Details}
\label{sec:appendix_tech_details}

\begin{figure}[t]
\centering
\includegraphics[width=0.9\linewidth]{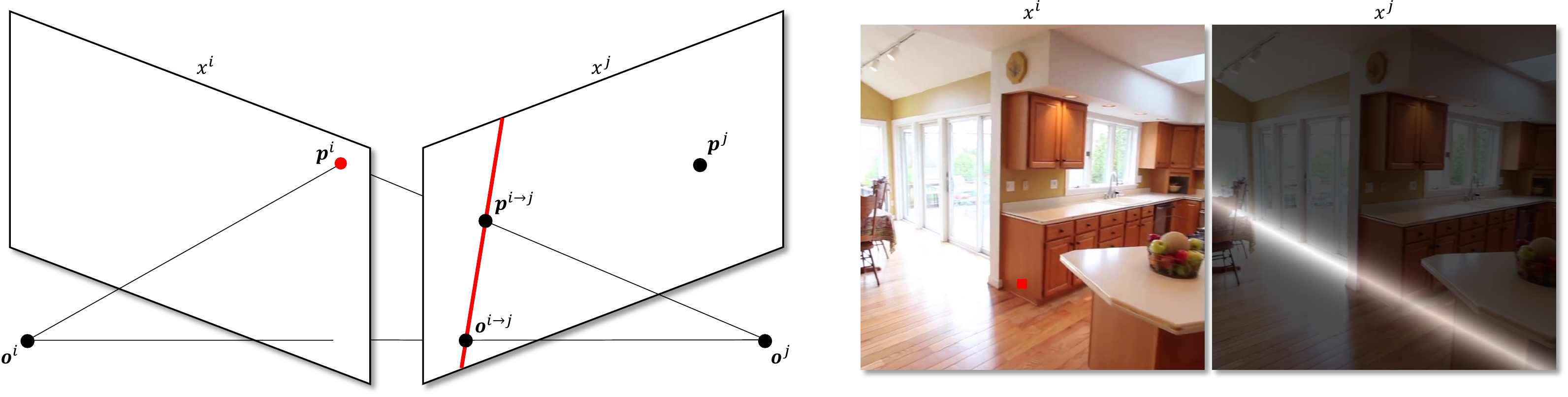}
\caption{\textbf{Epipolar line.} 
(\textit{left}) We show the computation of the epipolar line (red line) at the source view image $x^j$ which corresponds to the point $\mathbf{p}^i$ (red dot) on the target view image $x^i$.
(\textit{right}) 
We visualize the epipolar line on the source view image $x^j$ of the point $\mathbf{p}^i$ (red dot) on the target view image $x^i$.
We compute the weight map (right image) according to the epipolar line using \eqnref{weight_map}.
}
\label{fig:epipolar_line}
\end{figure}

\Paragraph{Epipolar line computation.}
We present the epipolar line computation in \figref{epipolar_line}.
Given a point $\mathbf{p}^i$ on the image plane at the target view $i$ and the relative camera pose $\{\mathbf{K}, \mathbf{R}^{i\rightarrow j}, \mathbf{t}^{i\rightarrow j}\}$, the goal is to find the corresponding epipolar line on the image plane at the source view $j$.
We first project the point $\mathbf{p}^i$ onto the source view image plane as $\mathbf{p}^{i\rightarrow j}$, namely
\begin{equation}
\mathbf{p}^{i\rightarrow j}=\pi\big(\mathbf{R}^{i\rightarrow j}(\mathbf{K}^{-1}\mathbf{p}^i) + \mathbf{t}^{i\rightarrow j}\big),
\end{equation}
where $\pi$ is the projection function.
We also project the camera origin $\mathbf{o}^i=[0,0,0]^T$ at the target view $i$ onto the source view image plane as $\mathbf{o}^{i\rightarrow j}$:
\begin{equation}
\mathbf{o}^{i\rightarrow j}=\pi\big(\mathbf{R}^{i\rightarrow j}(\mathbf{K}^{-1}[0,0,0]^T) + \mathbf{t}^{i\rightarrow j}\big).
\end{equation}
Then the epipolar line of the point $\mathbf{p}$ on the source view image plane can be formulated as
\begin{equation}
\mathbf{p}_\mathrm{epipolar} = \mathbf{o}^{i\rightarrow j} + c(\mathbf{p}^{i\rightarrow j} - \mathbf{o}^{i\rightarrow j})\hspace{5mm}c\in\{-\infty,\infty\}\in\mathbb{R}.
\end{equation}
Finally, the distance between a point $\mathbf{p}^j$ on the source view image plane and the epipolar line can be computed as
\begin{equation}
d(\mathbf{p}_\mathrm{epipolar}, \mathbf{p}^j)=\|(\mathbf{p}^j - \mathbf{o}^{i\rightarrow j})\times (\mathbf{p}^{i\rightarrow j}- \mathbf{o}^{i\rightarrow j})\| / \|\mathbf{p}^{i\rightarrow j}- \mathbf{o}^{i\rightarrow j}\|,
\end{equation}
where $\times$ and $\|\cdot\|$ indicate vector cross-product and vector norm, respectively.
According to the epipolar line, we compute the weight map, where higher pixel values indicate closer distance to the line
\begin{equation}
\label{eq:weight_map}
m_{\mathbf{p}^i,\mathbf{K}, \mathbf{R}^{i\rightarrow j}, \mathbf{t}^{i\rightarrow j}}(\mathbf{p}^j)= 1 - \mathrm{sigmoid}\big(50(d(\mathbf{p}_\mathrm{epipolar}, \mathbf{p}^j) - 0.05)\big)\hspace{5mm}\forall\mathbf{p}^j\in x^j.
\end{equation}
We use the constant $50$ to make the sigmoid function steep, and use the constant $0.05$ to include the pixels that are nearby the epipolar line.
An example weight map is visualized in the right-hand side of \figref{epipolar_line}.
After estimating the weight maps for all positions in the source view image, we stack these maps and reshape to get the epipolar weight matrix $E_{i,j}$, which is used to compute the epipolar attention described in (6) in the paper.
Note that if the epipolar line does not intersect with the target view, we assign the same value for all spatial positions in the epipolar weight matrix $E_{i,j}$.
This makes our epipolar attention falls back to the common cross-attention.

\Paragraph{Stochastic frame sampling.}
We randomly sample the source view in both the training and inference stages.
We observe inferior results if we use stochastic sampling only during the inference time.
We hypothesize this is due to the mismatch of the input relative pose distributions between the training and inference stages.

\begin{figure}[t]
\centering
\includegraphics[width=0.7\linewidth]{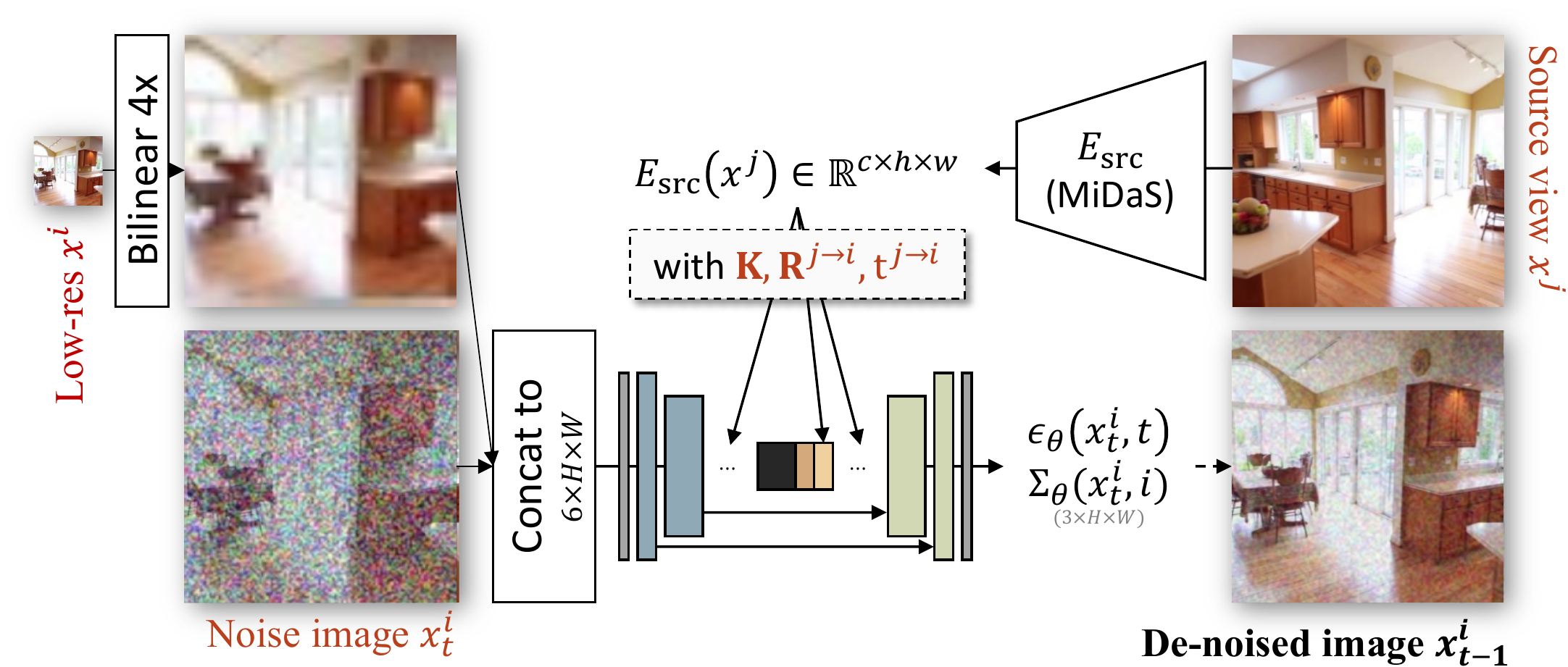}
\caption{\textbf{Super-resolution pose-guided diffusion model.} 
We first bilinearly up-sample the low-resolution target view, and concatenate it with the noised high-resolution target view. 
Combining the features extracted from the high-resolution source view, we train the UNet network to de-noise the high-resolution target view.
}
\label{fig:superres}
\end{figure}

\Paragraph{Super-resolution.}
We present the super-resolution model details in \figref{superres}.
We first bilinearly up-sample the low-resolution target view,
and concatenate it with the noised high-resolution target view. 
Combining the features extracted from the high-resolution source view, we train the UNet network to de-noise the high-resolution target view. 
We empirically find that taking the source view as input improves the temporal consistency in high-resolution videos.


\begin{figure}[t]
\centering
\includegraphics[width=0.9\linewidth]{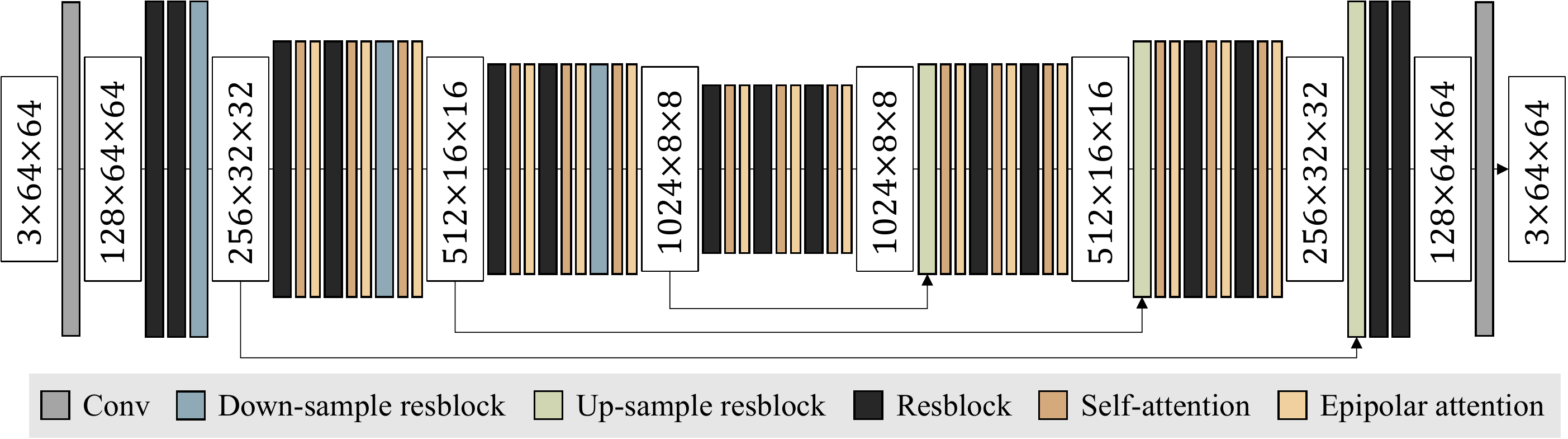}
\caption{\textbf{UNet architecture.} 
We present the UNet architecture of the pose-guided diffusion model that produces $64\times 64$ novel views.
}
\label{fig:unet}
\end{figure}

\Paragraph{Training details.}
\begin{table}[t]
    \caption{\textbf{Training details.} We provide the details of the UNet architecture and the training hyper-parameters.}
    \label{tab:training_details}
    \centering
    \begin{tabular}{l|cc} 
	\toprule
        Models & $64$ & $64\rightarrow256$ \\
        \midrule
        Channels & $128$ & $128$ \\
        Number of residual blocks & $3$ & $3$ \\
        Channel multiples & $1,2,3,4$ & $1,1,2,3,4$\\
        Head channels & $64$ & $64$ \\
        Attention resolutions (self and epipolar) & $32,16,8$ & $16$ \\
        Dropout (training) & $0.1$ & $0.1$ \\
        \midrule
        Diffusion steps (training) & $1000$ & $1000$ \\
        Noise schedule & cosine & cosine \\
        Sampling steps (inference) & $250$ & $250$ \\
        Sampling variance method & learned~\cite{improvedddmp} & DDPM~\cite{ddpm} \\
        \midrule
        Batch size & $64$ & $32$ \\
        Iterations & $1$M & $1$M \\
        Learning rate & $0.0001$ & $0.0001$ \\
        Adam $\beta_2$ & $0.999$ & $0.999$ \\
        Adam $\epsilon$ & $1$e-$8$ & $1$e-$8$ \\
        EMA decay & $0.9999$ & $0.9999$ \\
        \bottomrule
        
    \end{tabular}
\end{table}
We provide the training details, including the UNet architecture as well as the hyper-parameters in \figref{unet} and \tabref{training_details}.
We implement the proposed method with PyTorch~\cite{paszke2017automatic}, and use $8$ Nvidia V100 GPUs to conduct the training.

\subsection{Experiment Details}
\label{sec:appendix_exp_details}

\Paragraph{Dataset processing.}
We follow Lai~\etal~\cite{videoautoencoder} to process the RealEstate10K (Re10K)~\cite{re10k} and Matterport3D (MP3D)~\cite{mp3d} datasets and split them into training set and test set. 
Note that the availability of the YouTube videos in the Re10K dataset changes over time. 
From the available videos, we remove the short videos (less than 200 frames) and eventually obtain $61,986$ videos in the training set and $1,763$ videos in the testing set. 
We randomly select $500$ videos from the testing set for all quantitative and qualitative experiments.
As for the MP3D dataset, there are $6,097$ videos in the training set and $1,062$ videos in the test set, and again we select 500 videos from the test set for evaluation.

\Paragraph{Evaluation metrics.}
We use the AlexNet model to compute the LPIPS score, and InceptionV3~\cite{inceptionv3} network to calculate the FID/KID scores.\footnote{\url{https://github.com/richzhang/PerceptualSimilarity}}\footnote{\url{https://github.com/mseitzer/pytorch-fid}}\footnote{\url{https://github.com/GaParmar/clean-fid}}
Both the AlexNet and InceptionV3 models are pre-trained on the ImageNet dataset.
We use $30,000$ training images to pre-compute the real data statics for estimating the FID/KID scores.
On the other hand, we use the pre-trained RAFT~\cite{teed2020raft} model to compute the flow warping error $E_\mathrm{warp}$~\cite{flowwarpingerror}.\footnote{\url{https://github.com/princeton-vl/RAFT}}

\Paragraph{Reproducing results of previous methods.} We describe how we evaluate the previous methods as follows:
\begin{itemize}
\item GeoGPT~\cite{geogpt}: We use the official implementation.\footnote{\url{https://github.com/CompVis/geometry-free-view-synthesis}}
For the evaluation conducted on the Re10K dataset, we use the pre-trained model parameters released on the Github webpage.
Since the resolution of the generated images is slightly different from our setting, we center-crop and resize the images to $256\times 256$ for evaluation.
As for the MP3D dataset, we train the model using the default training parameters.

\item LoR~\cite{lor}:
We use the official implementation to conduct the training and evaluation.\footnote{\url{https://github.com/xrenaa/Look-Outside-Room}}
For the Re10K dataset, we train and evaluate the model using the default hyper-parameters.
As for the MP3D dataset, we use the pre-trained model parameters provided by the authors to conduct the evaluation.

\item SE3DS~\cite{se3ds}:
We use the official implementation and pre-trained model parameters to conduct the evaluation on the Re10K dataset.\footnote{\url{https://github.com/google-research/se3ds}}

\end{itemize}

\end{document}